\documentclass[lettersize,journal]{IEEEtran}
\usepackage{amsmath,amsfonts}
\usepackage{algorithmic}
\usepackage{algorithm}
\usepackage{array}
\usepackage{booktabs}
\usepackage[caption=false,font=normalsize,labelfont=sf,textfont=sf]{subfig}
\usepackage{color}
\usepackage{textcomp}
\usepackage{multirow}
\usepackage{ulem}
\usepackage{stfloats}
\usepackage{url}
\usepackage{hyperref}
\usepackage{verbatim}
\usepackage{graphicx}
\usepackage{marvosym}
\usepackage{cite}
\begin{document}

\title{Few-shot Object Localization}

\author{Yunhan Ren$^*$, Bo Li$^*$, Chengyang Zhang, Yong Zhang$\textsuperscript{\Letter}$, Baocai Yin \\ Beijing Key Laboratory of Multimedia and Intelligent Software Technology, Faculty of Information Technology, Beijing Institute of Artificial Intelligence, Beijing University of Technology, Beijing, 100124, China
        % <-this % stops a space
\thanks{$\textsuperscript{\Letter}$ Corresponding author: zhangyong2010@bjut.edu.cn. $^*$ Yunhan Ren and Bo Li contributed equally to this manuscript.}}

% The paper headers
\markboth{}%
% \markboth{IEEE TRANSACTIONS}%
{Shell \MakeLowercase{\textit{et al.}}: A Sample Article Using IEEEtran.cls for IEEE Journals}

\maketitle

% Existing few-shot object counting tasks primarily focus on quantifying the number of objects in an image, neglecting precise positional information. 
\begin{abstract}
Existing object localization methods are tailored to locate specific classes of objects, relying heavily on abundant labeled data for model optimization. However, acquiring large amounts of labeled data is challenging in many real-world scenarios, significantly limiting the broader application of localization models. To bridge this research gap, this paper defines a novel task named Few-Shot Object Localization (FSOL), which aims to achieve precise localization with limited samples. This task achieves generalized object localization by leveraging a small number of labeled support samples to query the positional information of objects within corresponding images. To advance this field, we design an innovative high-performance baseline model. This model integrates a dual-path feature augmentation module to enhance shape association and gradient differences between supports and query images, alongside a self query module to explore the association between feature maps and query images. Experimental results demonstrate a significant performance improvement of our approach in the FSOL task, establishing an efficient benchmark for further research. All codes and data are available at \href{https://github.com/Ryh1218/FSOL}{https://github.com/Ryh1218/FSOL}.

\end{abstract}


\section{Introduction}
\IEEEPARstart{O}{bject} localization \cite{OLOverall1, OLOverall2, TMM4}, a fundamental task in computer vision, has witnessed remarkable progress propelled by deep learning techniques. Precise object localization within images is critically important across various applications, including autonomous vehicles \cite{AutoCar}, surveillance systems \cite{OLSS}, medical image analysis \cite{li2024multi, OLMI}, and crowd management \cite{TMM5}. Despite significant advancements, existing methods predominantly rely on abundant labeled data to train high-accuracy models. However, acquiring such labeled datasets in real-world scenarios often poses formidable challenges due to associated expenses and time constraints. In response to these challenges, few-shot learning has emerged as a promising paradigm to reduce dependence on extensive labeled datasets \cite{FSLOverall11, FSLOverall12, FSLOverall2}. By enabling models to learn from a limited number of labeled examples, few-shot learning enhances generalization capabilities, proving particularly advantageous in scenarios where substantial labeled data is impractical or unfeasible.

\begin{figure*}[!tp]
\centering
{\includegraphics[width=7in]{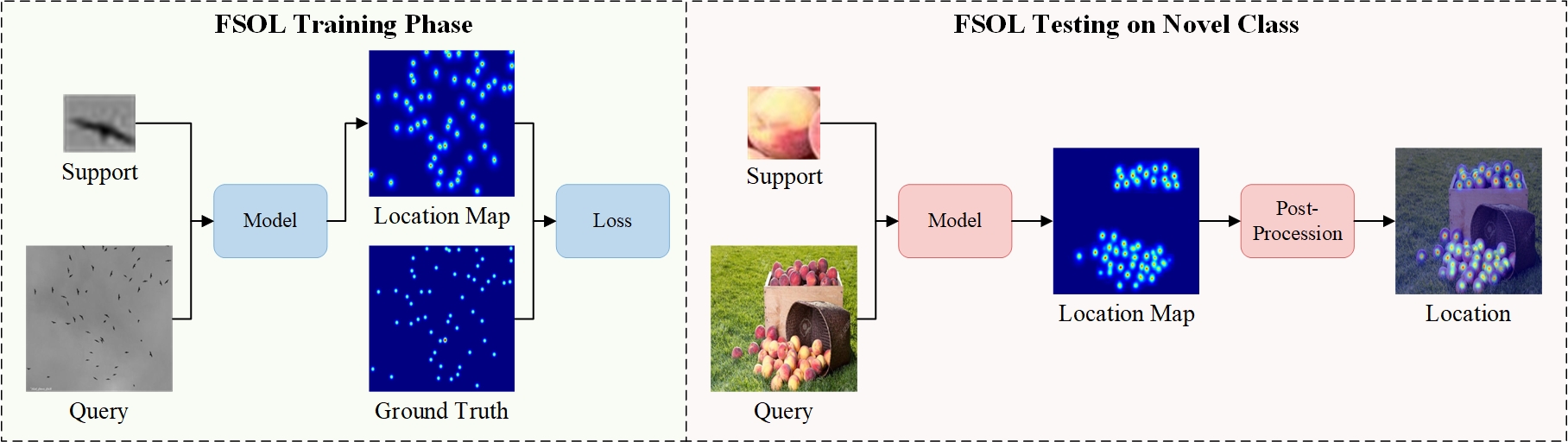}%
\caption{Demonstration of the Few-Shot Object Localization (FSOL) task. During the training phase, the model predicts the location map based on given support samples and their corresponding query image. It then adjusts its parameters by by minimizing the Mean Squared Error loss between the ground truth and the predicted location map. In the testing phase, the trained model predicts the location map of novel class samples on corresponding query images that were not appear in the training phase.}
\label{intro-task}}
\end{figure*}

With the advances in few-shot learning, accurate object localization without a large amount of labeled data is achievable yet unexplored. Consequently, we further explore object localization under few-shot settings, namely, when given an image with only a few samples labeled, locating other same-category samples solely with the assistance of the labeled ones. We term this problem the Few-Shot Object Localization (FSOL) task. Unlike few-shot object counting tasks \cite{FSCTask} that primarily focus on quantity analysis, FSOL emphasizes recognizing objects and providing accurate positional information within images. As illustrated in Fig. \ref{intro-task}, the model learns from labeled support samples of known categories during the training phase. In the testing phase, the model demonstrates a remarkable ability to generalize to new categories, significantly enhancing its overall adaptability and performance.

\begin{figure}[tp!]
        \center
        \scriptsize
        \begin{tabular}{cc}
                \includegraphics[width=3.2in]{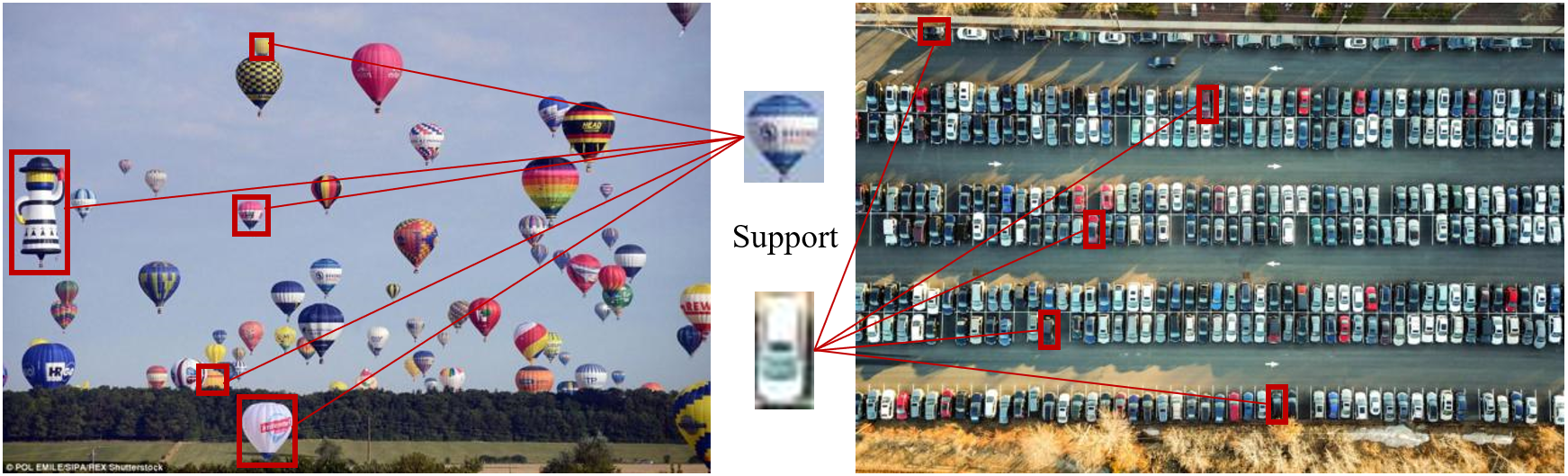} \\ (a) Appearance gap between intra-class objects \\
                \includegraphics[width=3.2in]{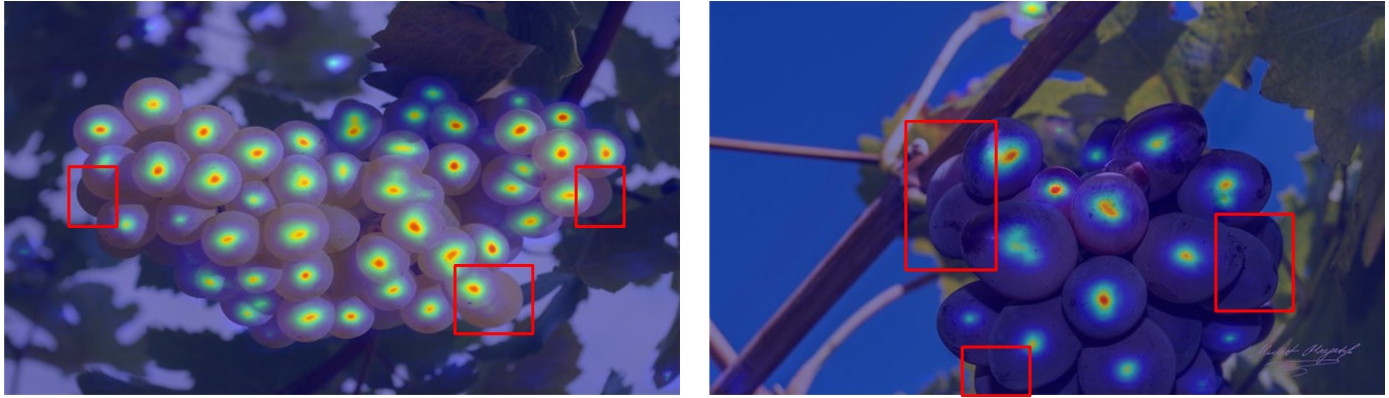} \\
                (b) Object omission due to inter-object occlusion   \\
        \end{tabular}
        \caption{Difficulties in few-shot object localization that cause negative influences: a) Appearance gap between intra-class objects; b) Object omission due to inter-object occlusion.}
        \vspace{-0.5em}
\label{intro}
\end{figure}

This study aims to advance the research field by introducing a high-performance benchmark model for the FSOL task. Within this task, we have identified two primary challenges: 1) Appearance gap between intra-class objects: Significant variations among objects within the same class in the query image create an appearance gap compared to the support image samples, impacting query accuracy (see Fig. \ref{intro}(a)). 2) Object omission due to inter-object occlusion: The model struggles to accurately distinguish dense, overlapping objects in the query image, leading to a decrease in localization recall rate (see Fig. \ref{intro}(b)).

To tackle the key challenges in the FSOL task, we design a dual-path feature augmentation module. To handle intra-class variations in shape, size, and orientation, we use a deformable convolution branch \cite{DCv1, DCv2}, which enhances localization performance by adapting to feature variations. To reduce object omission, we implement a cross-center difference convolution branch \cite{CCD-C}, improving feature discriminability by capturing gradient difference. Additionally, we introduce 3D convolution to capture image structure, texture, and patterns, enhancing feature representation and model performance. The query image performs 3D convolution with the support image to generate a similarity map reflecting object locations.

Further, leveraging the original query image to enhance the obtained similarity map has emerged as a promising strategy. Current methods often add the query feature directly to the similarity feature map \cite{SafeCount}, using a residual connection technique to retain object information from the original image and optimize the similarity feature map. However, this direct addition strategy introduces considerable noise from the query image, making it unsuitable for localization tasks that demand high accuracy. Therefore, inspired by self-support matching \cite{SSP}, we utilize the similarity matrix calculated between the query image and the similarity map for weighting. This approach aims to more accurately incorporate the query image's information while reducing undesirable noise.

In this paper, we introduces the pioneering task of FSOL and presents an innovative high-performance benchmark. To tackle the challenges posed by significant intra-class variations and target occlusions in localization tasks, we design a dual-path feature augmentation module that aims to enhance appearance correspondence and gradient discrimination between support and query features. Moreover, to effectively leverage information in query images for enhancing similarity maps, we introduce a self query module to explore intricate associations between feature maps and query images. Experimental results demonstrate the substantial performance improvement of our approach in the FSOL task, establishing an efficient benchmark for further research in object localization under limited data scenarios. In summary, the contributions of this paper can be outlined as follows:
\begin{itemize}
\item{For the first time, we define the task of few-shot object localization, proposing a new research direction for object localization in scenarios with limited labeled data, and establish a high-performance benchmark.}
\item{The dual-path feature augmentation module is designed to simultaneously enhance the deformation and gradient association between support and query images, significantly improving localization performance.}
\item{The self-query module uses similarity matrix weighting to leverage the query image for enhancing the similarity map while avoiding excessive noise interference.}
\end{itemize}

\section{Related Work}
This paper is the first work to explore the task of few-shot object localization. Therefore, in this section, we briefly review the research in other fields related to this paper. First, we outline some classic works for the object localization task, then briefly introduce the development of few-shot learning, and finally introduce the currently prevalent works for few-shot object counting.

\subsection{Object Localization}
Object localization aims to precisely determine the position of objects in images or videos. Unlike object detection, which involves bounding box information, object localization focuses solely on object position. This localization process typically involves two stages: generating a location map through model prediction, followed by post-processing to extract object position details \cite{FIDTM, EDT, li2024lite}. The evolution from density maps \cite{ShangHaiTech, CSRNet, TMM3} to location maps \cite{iKNN, FIDTM, EDT} emphasizes accurate object localization and high-quality gradient data for model training. Traditional density maps struggle with precise object localization, prompting innovations like the IDT map \cite{iKNN}, designed to provide accurate object positions through distance flipping. Despite improvements, IDT maps often contain significant background noise. Addressing this, Liang et al. \cite{FIDTM} introduced the Focal Inverse Distance Transform (FIDT) map, offering enhanced gradient information. However, issues like rapid foreground decay and slow background decline persist. To address these, Li et al. \cite{EDT} proposed using exponential functions to moderate foreground decay and expedite background zeroing, resulting in more optimal location maps.

The aforementioned studies have significantly advanced the field of object localization, providing a strong foundation for further exploration. Despite these strides, current research heavily relies on extensive labeled data, leaving the limited sample scenario underexplored. This gap presents an opportunity for future studies to delve into object localization within limited sample settings.

\subsection{Few-shot Learning}
Few-shot Learning (FSL) tackles the challenge of learning from a limited number of training samples \cite{FSLOverall2, TMM1, TMM2}. Unlike conventional deep learning tasks, FSL grapples with data scarcity, where supervision information is restricted, hindering the capture of diverse categories and variations. Typically, FSL research encompasses two primary stages: training models with sparse samples and then achieving generalization through fine-tuning with a small dataset. In the realm of few-shot image classification, Hu et al. \cite{ICBest} demonstrated exceptional performance leveraging meta-learning, pre-training, and fine-tuning strategies. Meanwhile, for few-shot object detection, Liu et al. \cite{ODBest} significantly enhanced average precision by adopting a pre-trained transformer decoder as the detection head and omitting the Feature Pyramid Network (FPN \cite{FPN}) in feature extraction. In few-shot semantic segmentation, Lu et al. \cite{SSBest} focused on optimizing the classifier via meta-learning, while Zhang et al. \cite{Pixel} achieved pixel-level feature alignment by integrating pixel-level support features into the query. These approaches present innovative solutions across various domains, highlighting FSL's prowess in addressing data scarcity challenges.

Few-shot learning methods offer competitive performance even with limited annotated data, a highly promising attribute. This paper proposes integrating few-shot learning techniques with object localization tasks to explore innovative methods for improving object localization performance in few-shot scenarios.

\subsection{Few-shot Object Counting}
The Few-shot Object Counting task aims to describe object categories using support images and then count the instances of these categories in the query image. Previous research \cite{zhang2024crowdgraph} primarily focused on enumerating specific objects within single categories, constraining the task's generalization capability. Lu et al. \cite{CC} introduced class-agnostic counting, developing a GMN model capable of counting objects across any category by reframing the problem as an image self-similarity matching task. In contrast, Yang et al. \cite{CFOCNet} employed multi-scale computation to assess similarity between support and query images, transforming object counting into a similarity matching challenge. Due to the absence of dedicated few-shot object counting datasets, Ranjan et al. \cite{FSC147} introduced the FSC-147 dataset, which includes 147 object categories and 6000 images, and proposed the FamNet model based on similarity comparison techniques. More recently, You et al. \cite{SafeCount} integrated feature enhancement and similarity matching concepts, achieving state-of-the-art performance in few-shot object counting. These studies contribute both theoretically and empirically to advancing the few-shot object counting task.

These studies indeed furnish a robust theoretical underpinning and empirical validation for the few-shot object counting task. However, a sole reliance on counting methods imposes certain limitations, especially in scenarios requiring a deeper understanding of object relationships. For instance, in medical image analysis \cite{MIA}, it's imperative to consider the mutual influence between disease features. Similarly, in autonomous driving systems, comprehending the intricate behaviors and interactions of traffic participants is paramount \cite{liu2021vehicle}. These considerations underscore the need to augment few-shot object counting with richer information beyond mere object quantity statistics.

\section{Method}
\begin{figure*}[!tp]
\centering
{\includegraphics[width=7in]{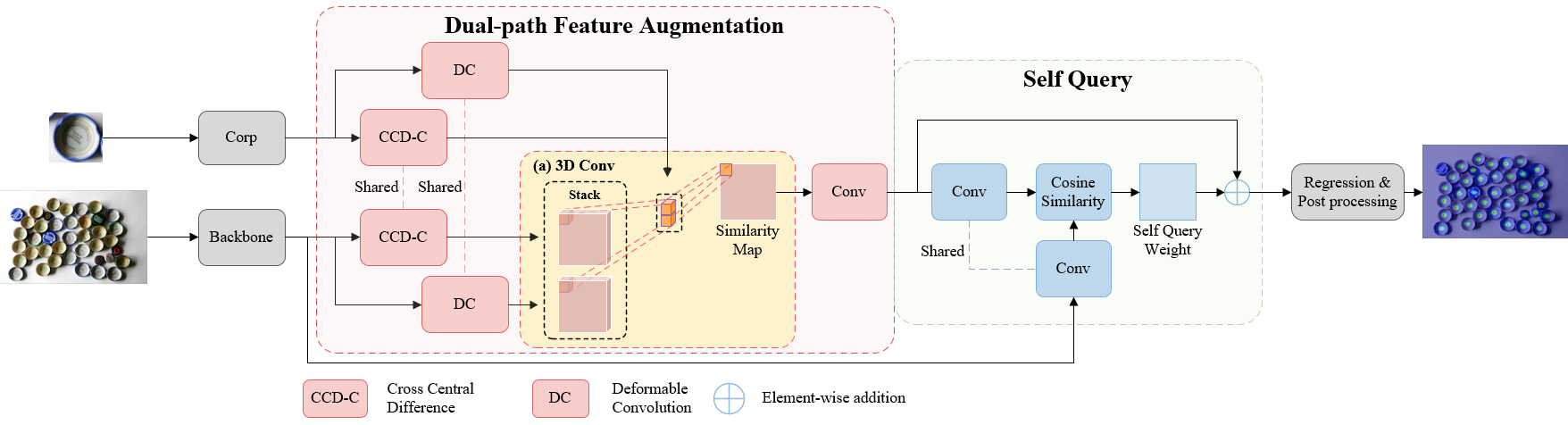}%
\caption{Demonstration of our FSOL Pipeline. Given the query and support images, the query feature $F_Q$ is extracted from the query image while the support feature $F_S$ is cropped from $F_Q$. The Dual-path Feature Augmentation (DFA) module first enhances deformation and gradient information in both $F_Q$ and $F_S$ through deformation and gradient branches, outputting the deformation-enhanced $F_Q^D$, $F_S^D$ as well as gradient-enhanced $F_Q^C$ and $F_S^C$. Then, DFA performs 3D convolution on the stacked $F_Q^D$ and $F_S^D$ using stacked $F_Q^C$ and $F_S^C$ as convolution kernel weights, obtaining the similarity map $S$ between query and support images. Then, the Self Query (SQ) module accepts $S$ as input and uses the original $F_Q$ to guide the object’s distribution information in $S$, subsequently outputting the optimized similarity map $S_{SQ}$. The $S_{SQ}$ then be sent to regression head to get the final location map.}
\label{pipeline}}
\end{figure*}

This study proposes a novel Few-Shot Object Localization (FSOL) task to enhance object localization accuracy and generalization in few-shot scenarios. To this end, we employ three key strategies, as shown in Fig. \ref{pipeline}. Firstly, our method emphasizes capturing deformation and gradient information within the image to better understand the association between query and support features. Secondly, we integrate 3D convolution to comprehensively consider various channels' impact on feature similarity comparison. Lastly, we introduce a module utilizing original query features to guide similarity distribution, allowing the model to utilize optimized similarity information for identifying and locating challenging samples. This reduces object omissions and enhances the model's generalization. 

\subsection{Dual-path Feature Augmentation}
In this study, we introduce a Dual-path Feature Augmentation (DFA) module aimed at mitigating significant intra-class appearance gaps and addressing object omission challenges in FSOL task. The DFA module comprises two key components: a deformation branch and a gradient branch. The deformation branch facilitates the model to learn the matching relationships within intra-class features, thereby improving localization accuracy by capturing deformable features in both support and query images. On the other hand, the gradient branch emphasizes comparing gradient similarities between support and query images during similarity computation. This enhances the model's ability to discern object features, thereby effectively reducing object omissions. The subsequent sections will elaborate on the vanilla 2D convolution, followed by a detailed explanation of the deformation branch and gradient branch.

\subsubsection{Vanilla 2D Convolution}
In convolutional neural networks, the vanilla 2D convolution serves as a fundamental operator, comprising two primary steps: (i) Sampling local neighboring regions on the input feature map $x$ using $N$ sampling points; (ii) Aggregating the sampled values through learnable weights $w$. Therefore, for the position $p$ on the input feature map $x$, the corresponding value at the same position $p$ on the output feature map $y$ can be mathematically expressed as:
\begin{equation}
\label{vanilla_conv}
y\left(p\right)=\sum_{k=1}^{N}{w_k \cdot x\left(p+p_k\right)}\
\end{equation}
where $p_k$ represents the predefined offset of the enumerated position $p$ in the local neighbor area. For example, when $N=9$, there are nine sampling points, $p_k=\{(-1,-1),(-1,0),\ldots,(0,1),(1,1)\}$. In this paper, we initially apply vanilla 2D convolution to process the image. For an input image $\mathbb{R}^{1\times3\times512\times512}$, we use the first three stage outputs of ResNet50, resulting in outputs of 256, 512, and 1024 channels, respectively. These outputs are concatenated into a feature tensor $F\in\mathbb{R}^{1\times1792\times128\times128}$. Subsequently, query features $F_Q\in\mathbb{R}^{1\times256\times128\times128}$ are obtained through the aforementioned vanilla 2D convolution. For our primary experiment, the support feature 
$F_S\in\mathbb{R}^{1\times256\times x\times y}$ is cropped from the query feature $F_Q$ based on the dimensions of the original input sample. Following an adaptive pooling layer, the support feature is resized to $F_S\in\mathbb{R}^{1\times256\times3\times3}$. Subsequently, these features are forwarded to both the deformation branch and the gradient branch for further processing.

\subsubsection{Deformation Branch}
In FSOL task, the object within the query image often exhibits substantial intra-class variation, distinct from the object depicted in the support image, thereby resulting in inaccurate queries. To address this issue, we aim to enhance the model's capability to capture deformable information present in both query and support features. 

\begin{figure*}[!t]
\centering
\includegraphics[width=6.4in]{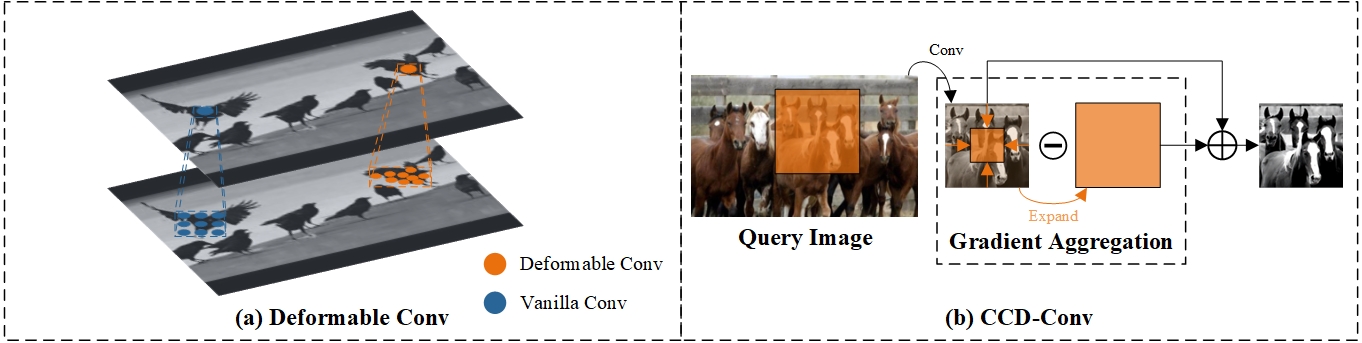}
\caption{Demonstration of two convolutional strategies for enhancing deformation and gradient information in support and query images: (a) Deformable Conv: Vanilla convolution utilizes fixed sampling points, potentially introducing noise, while deformable convolution adjusts sampling points, reducing background noise and improving adaptability; (b) CCD-Conv: Cross-center difference convolution computes differences between neighboring pixels around the central pixel and uses these differences as weights to generate the final output. This approach captures subtle image changes like texture, edges, and fine details.}
\label{intro-conv}
\end{figure*}

To this end, we incorporate a Deformable Convolution (DC \cite{DCv1, DCv2}) branch. Unlike traditional convolution, which relies on fixed sampling points within the receptive field, DC introduces an offset to each sampling point, thereby making their positions variable. This offset is generated by another convolution layer. In Fig. \ref{intro-conv}(a), we depict the differences between sampling points in vanilla and deformable convolution. While vanilla convolution tends to integrate redundant information into features, deformable convolution diminishes redundancy by iteratively optimizing sampling points. To assess the relevance of the introduced area, DC incorporates an additional weight coefficient. When the area is deemed irrelevant, this coefficient is set to 0. Mathematically, for the same position $p$ in the input feature map $x$ and the output feature map $y$, the relationship between the values $x(p)$ and $y(p)$ after deformable convolution can be expressed as follows:
\begin{equation}
\label{d_c}
 y(p) = \sum_{k=1}^{N} w_k \cdot x(p + p_k + \Delta p_k) \cdot \Delta m_k \quad
\end{equation}
where $\Delta p_k$ represents the offset learned by each sampling point, and $\Delta m_k$ represents the weight coefficient in the range [0, 1]. In our framework, both the original query and support features fed into deformation branch are processed by a shared DC layer to highlight their deformation information. Specifically, the query feature $F_Q\in\mathbb{R}^{1\times256\times128\times128}$ and support feature $F_S\in\mathbb{R}^{1\times256\times3\times3}$ each go through a deformable convolution layer, resulting in the enhanced query feature $F_Q^D\in\mathbb{R}^{1\times256\times128\times128}$ and the corresponding enhanced support feature $F_S^D\in\mathbb{R}^{1\times256\times3\times3}$. This design not only reduces the amount of computation but also unifies the deformation distribution of both query and support features, alleviating the appearance gap between intra-class samples.

\subsubsection{Gradient Branch}
In FSOL task, the overlap of samples, unclear edges, and brightness differences also lead to inaccurate results when the model detects and locates these samples. For this reason, we introduce a gradient path to enrich the gradient information of support and query features, thereby helping the model better understand the edge and texture information of samples in the image.

Specifically, we choose the Cross Central Difference Convolution (CCD-C) technique to decouple the central gradient features into cross-directional combinations, thereby aggregating messages more effectively and with greater focus \cite{CCD-C}. Compared with vanilla convolution, CCD-C is more sensitive to the edges and textures of objects. This sensitivity improves the model’s ability to better distinguishing unique samples, thereby improving its localization accuracy. At the same time, compared with aggregating the vanilla gradient features and central gradient features of the entire local neighboring area $R$, we choose to aggregate the gradient information only in the horizontal and vertical directions to reduce feature redundancy. As shown in Fig. \ref{intro-conv}(b), the above process is represented by the following formula:
\begin{equation}
\label{ccd_c}
\begin{split}
    y(p) = \theta \cdot \sum_{k=1}^{N_{HV}} w_k \cdot (x(p + p_k) - x(p)) \\ +  (1 - \theta) \cdot \sum_{k=1}^{N_{HV}} w_k \cdot x(p + p_k) \quad
\end{split}
\end{equation}
where $p_k\in\{(-1,0),(0,-1),(0,0),(0,1),(1,0)\}$, since only the gradient information in the horizontal and vertical directions is aggregated. $\theta\in[0,1]$ is used to balance the intensity-level and gradient-level information.

In our framework, both the query and support features are processed by a single CCD-C(HV) layer with shared parameters. Specifically, the query feature $F_Q\in\mathbb{R}^{1\times256\times128\times128}$ and support feature $F_S\in\mathbb{R}^{1\times256\times3\times3}$ are optimized to enhanced features $F_Q^C\in\mathbb{R}^{1\times256\times128\times128}$ and $F_S^C\in\mathbb{R}^{1\times256\times3\times3}$ after being processed by CCD-C layer individually. This approach not only reduces computational cost but also boosts the model’s sensitivity to gradient information, such as edges and brightness. In essence, it enables the model to detect boundaries and brightness changes more effectively across different samples, thereby enhancing accuracy in object identification and categorization. This serves to alleviate the issue of object omission triggered by gradient-related factors such as brightness and edges.

\subsection{3D Convolution}
After processed by the aforementioned two branches, both the support and query features are enriched with gradient and deformation information. This enhancement equips our model to efficiently and accurately learn the rich yet noisy information encapsulated within these features. However, enhancing them separately overlooks the holistic nature of the features and fails to leverage the inherent relationships between the information they contain. To tackle these issues, we leverage 3D convolution to integrate information from both branches and facilitate comparison between the support and query features. 3D convolution allows for simultaneous processing of information across multiple channels by introducing a depth dimension. Thus, for a given position $p$ on the input feature map $x$, the corresponding value at the same position $p$ on the output feature map $y$ can be mathematically expressed as:
\begin{equation}
\label{3d_c}
 y(p) = \sum_{d=1}^{D} \sum_{k=1}^{N} w_{kd} \cdot x(p + p_{kd})
\end{equation}
where $D$ represents the depth dimension of the input feature map, $p_k\in\{(-1,-1,-1),(-1,-1,0),\ldots,(0,1,1),(1,1,1)\\\}$. We concatenate the outputs of the two branches along different dimensions, effectively creating a multi-dimensional feature space. This concatenated feature is then subjected to a 3D convolutional operation. Here, the support features are sequentially stacked and used as the convolution kernel, convolving with the similarly stacked query features. Throughout this process, the convolution along the depth dimension integrates both deformation and gradient information, thereby generating an association map between the support and query features. \textcolor{black}{In other words, enhanced features in different aspects can be aggregated together to yield a more comprehensive and accurate feature representation.}

Specifically, the model inputs the DC-enhanced query feature $F_Q^D\in\mathbb{R}^{1\times256\times128\times128}$ and support feature $F_S^D\in\mathbb{R}^{1\times256\times3\times3}$, as well as CCD-C-enhanced features $F_Q^C\in\mathbb{R}^{1\times256\times128\times128}$ and $F_S^C\in\mathbb{R}^{1\times256\times3\times3}$. Subsequently, the 3D convolution employs the parameters of the support features $F_S^D$ and $F_S^C$ as convolution kernel weights. It then conducts convolution in the $H$, $W$, and depth $D$ directions, corresponding to the stacking direction of $F_Q^D$ and $F_Q^C$, finally obtaining the preliminary similarity map $S\in\mathbb{R}^{1\times256\times128\times128}$. Since the stacking order in the depth direction is specified during input, the support features outputted by the two branches convolve on the corresponding query feature without interfering with each other. This process can be represented by the following formula:
\begin{equation}
\label{3d_o}
 y(p) = \sum_{i}^{Z} \sum_{k=1}^{N} w_{i} \cdot x(p + p_{ki})
\end{equation}
where $Z \in\{F_S^D,\ F_S^C\}$, $w_i$ represents the kernel weights of $F_S^D$ and $F_S^C$, $p_{ki}$ represents the sampling window area on query features $F_Q^D$ and $F_Q^C$. Consequently, the similarity map $S$ obtained through 3D convolution effectively integrates the deformation and gradient information of the enhanced features, thereby improving the model’s ability to represent features.

\subsection{Self Query Module}
The DFA module and 3D convolution significantly enhance the model's feature extraction capability. However, they solely focus on mining the similarity between support samples and query images to generate a similarity map, disregarding the distribution information of objects in the query image. To overcome this limitation, inspired by the self-support matching \cite{SSP}, we design the Self Query (SQ) module to integrate the distribution information of the query image into the obtained similarity map. This integrated information guides the optimization process, rather than relying solely on a blind search for similar samples in the similarity map. Essentially, the model gradually learns to utilize the information from the query image to steer optimization of the obtained feature comparison results. As a result, it achieves high precision in feature matching even with limited information from support samples.

\begin{figure*}[]
\centering
{\includegraphics[width=7in]{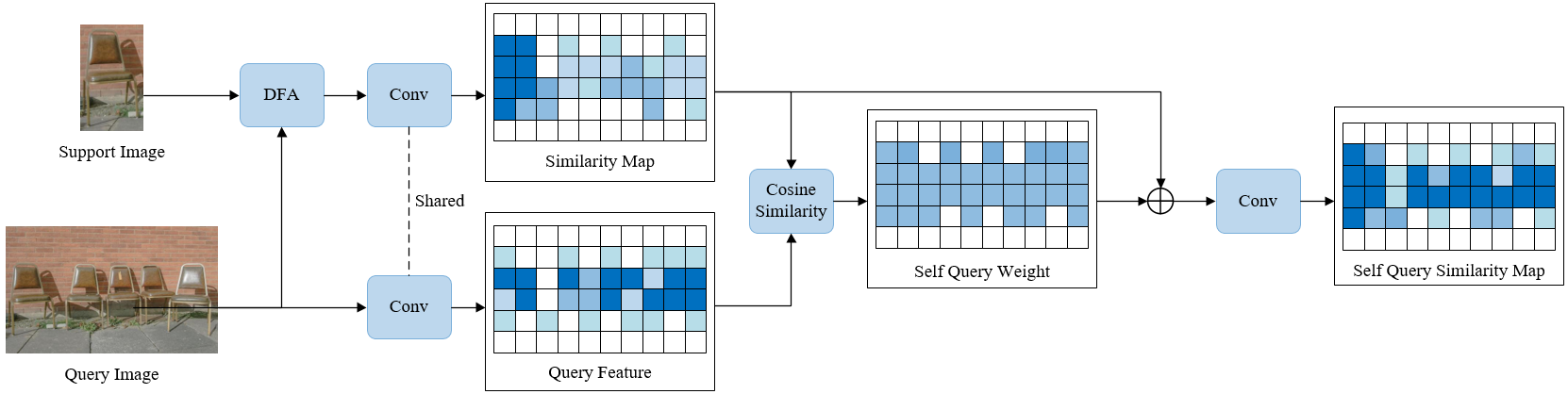}%
\caption{Demonstration of Self Query (SQ) module. The SQ module enhances the model's perception of the object distribution by integrating information from the similarity map $S$ and the original query image features $F_Q$. Initially, it applies a shared convolution layer to both $S$ and $F_Q$, thereby introducing non-linearity and capturing similar patterns. Next, it calculates the cosine similarity between $S$ and $F_Q$ to obtain self query weights $W$. These weights are added element-wise to $S$, enabling the distribution information from $S$ to guide optimization through $F_Q$. Finally, after passing through another convolution layer, the SQ module generates the optimized similarity map $S_{SQ}$.}
\label{sq_pipeline}}
\end{figure*}

The SQ module enhances model's perception of the object distribution by integrating query image features and the similarity map through several key steps.  Initially, it employs two 1$\times$1 convolution layers with shared parameters to introduce non-linearity and capture similar patterns in both the original query features and the obtained similarity map. Following the computation of cosine similarity between the query features and the similarity map, self query weights are derived. These weights are then added element-wise to the original query features, allowing the model to leverage the distribution information of objects in the query image for optimizing the similarity map. This iterative process enables the model to prioritize relevant regions within both the query features and the similarity map. Subsequently, the resulting similarity is employed as weights to refine the similarity map, enhancing the clarity of foreground objects while maintaining smoothness in the background. This refined representation enables the similarity map to more accurately depict support samples present on the query image. The comprehensive process is illustrated in Fig. \ref{sq_pipeline}. The aforementioned process can be represented as:
\begin{equation}
\label{sq_1}
 W=cosine\_similarity(In\_conv(S),\ In\_conv(F_q))
\end{equation}
\begin{equation}
\label{sq_2}
S_{sq}=Out\_conv(W+F_q)
\end{equation}
where $In\_Conv$ and $Out\_Conv$ represents 1$\times$1 convolution, $S\in\mathbb{R}^{1\times256\times128\times128}$ and $F_Q\in\mathbb{R}^{1\times256\times128\times128}$ represents the similarity map outputted by 3D convolution and original feature extracted from query image respectively. They are both inputted to a shared convolution layer and calculated a similarity weight $W\in\mathbb{R}^{1\times128\times128}$ by cosine similarity. Since the shape of the similarity map $S$ and similarity weight $W$ is same, thus they can directly perform element-wise addition and obtain the SQ-enhanced similarity map $S_{SQ}^T\in\mathbb{R}^{1\times256\times128\times128}$. Then, $S_{SQ}^T$ is processed by another 1$\times$1 convolution layer to get final enhanced similarity map $S_{SQ}\in\mathbb{R}^{1\times256\times128\times128}$ as the output of our SQ module. 
\begin{algorithm}
	%\textsl{}\setstretch{1.8}
	\renewcommand{\algorithmicrequire}{\textbf{Input:}}
	\renewcommand{\algorithmicensure}{\textbf{Output:}}
	\caption{Self Query Algorithm}
	\label{alg1}
	\begin{algorithmic}[1]
		\REQUIRE Query feature $F_{Q}$, Similar map $S$
		\STATE \textbf{function} $Self \ \ Query(F_{Q}, S)$
        \STATE $F_{Q}^{T} = In\_conv(F_{Q})$
        \STATE $S^{T} = In\_conv(S)$
        \STATE $W = Similarity(F_{Q}^{T}, S^{T})$
        \STATE $S_{SQ}^{T} = W + F_{Q}^{T}$
        \STATE $S_{SQ} = Out\_conv(S_{SQ}^{T})$
		\ENSURE  Self Query similarity map $S_{SQ}$
	\end{algorithmic}  
\end{algorithm}

To be more specific, we present the process using Algorithm \ref{alg1}. Here, $F_Q^T$, $S^T$ and $S_{SQ}^T$ denote the intermediate variables of $F_Q$, $S$ and $S_{SQ}$, respectively. The initial similarity map $S$, obtained through 3D convolution, is sensitized by the distribution information from the query image. This adaptation enables $S$ to more accurately represent the distinctive features present in the query image. Consequently, it becomes adept at identifying complex objects, maintaining robust localization capabilities even in scenarios where support information is limited.

\subsection{Regression and Post Processing}
After obtaining the optimized similarity map, the model undergoes regression and post-processing steps. The regression process involves upsampling the feature map to match the size of the input image, facilitating subsequent loss computation or further post-processing to derive coordinate information. This regression step is achieved through a series of convolutional layers, followed by Leaky ReLU activation and bilinear upsampling layers, drawing inspiration from designs presented in \cite{CC, FSC147, FasterRCNN}. Consequently, a location map of identical dimensions to the original image is generated. During training, the location map is directly compared with ground truth to compute the loss, subsequently updating the model parameters. During testing, the location map undergoes post-processing to extract target position information.

Specifically, the post-processing is implemented based on the Local-Maxima-Detection-Strategy described in FIDTM ~\cite{FIDTM}. Given a series of $M$ candidate points, only points with values greater than the threshold $T_a=100/255$ are selected as positive samples to filter out false positives. In FSOL tasks, object sizes vary significantly across different datasets. Therefore, we categorize datasets into dense and sparse datasets based on the average sample sizes. For dense datasets, the threshold is set to $T_a=40/255$. However, for sparse datasets, where object sizes tend to be larger compared to dense datasets, we adjust this threshold to $T_a=60/255$. Additionally, we enforce a lower limit of 0.06 for all candidate point values to filter out noise, implying that candidate points with values less than 0.06 are considered negative samples.

\section{Experiments}
\subsection{Datasets}
To validate the effectiveness of our method and demonstrate the high performance of the baseline framework, this paper primarily conducts experiments on four datasets. These datasets include dedicated few-shot datasets as well as densely populated datasets of crowds and vehicles.
\subsubsection{FSC-147}
The FSC-147 dataset is a unique collection of images tailored for the task of few-shot counting \cite{FSC147}. It features a diverse array of 147 distinct object categories, spanning from everyday items like kitchen utensils and office stationery to more complex entities such as vehicles and animals. This dataset comprises 6135 images, with the number of objects in each image varying significantly, ranging from as few as 7 to as many as 3731 objects. Each object instance within an image is denoted by a dot at its approximate center. Furthermore, for each image, three object instances are randomly selected as exemplar instances and annotated with axis-aligned bounding boxes. 
\subsubsection{ShangHaiTech}
The ShangHaiTech dataset is a comprehensive crowd counting dataset comprising two parts: ShangHaiTech Part A and ShangHaiTech Part B \cite{ShangHaiTech}. ShangHaiTech Part A consists of 482 images sourced from the internet. These images exhibit diversity and encompass a wide range of scenarios, rendering them suitable for training robust crowd counting models. The dataset is further partitioned into a training subset comprising 300 images and a testing subset comprising 182 images. 

Conversely, ShangHaiTech Part B comprises 716 images captured on the bustling streets of Shanghai. This segment of the dataset offers a realistic depiction of real-world crowd scenarios. It is divided into a training subset with 400 images and a testing subset with 316 images. Each image in both parts of the dataset is annotated with a dot marking the center of each person's head, providing precise ground truth labels for crowd counting. In total, the dataset contains annotations for 330,165 individuals.

\subsubsection{CARPK}
The car parking lot dataset (CARPK) is tailored for object counting and localization tasks, focusing specifically on counting cars in parking lots \cite{CARPK}. This dataset comprises 1448 images featuring nearly 90,000 cars, with each car annotated by a bounding box. The training set consists of three scenes, while an additional scene is reserved for testing purposes.

\subsection{Metrics}
\subsubsection{Localization}
This paper quantifies the localization error using the Euclidean distance between the center point of the predicted object and the corresponding ground truth point. Mathematically, it can be expressed as:
\begin{equation}
\label{l_error}
error=\sqrt{(x_{pred}-x_{gt})^2+(y_{pred}-y_{gt})^2}
\end{equation}
where $x_{pred}$ and $y_{pred}$ represent the $x$ and $y$ coordinates of predicted point, $x_{gt}$ and $y_{gt}$ represent the x and y coordinates of the ground truth point.

To facilitate a comprehensive comparison of localization performance among different models, a threshold $\sigma$ is established, representing the maximum allowable error. For each predicted point, its localization error with all ground truth points is computed. If the minimum error is less than or equal to $\sigma$, the predicted point is deemed a positive sample, indicating correct object localization by the model. Otherwise, if the minimum error exceeds $\sigma$, the predicted point is classified as a negative sample, signifying an incorrect object localization by the model. Given the wide range of sample sizes in FSOL, this paper predominantly utilizes $\sigma_l=10$ as the high threshold to assess model performance. However, on specific datasets, $\sigma_l=8$ is adopted to align with other model results. Furthermore, this paper presents results using 4 and 5 as the low threshold $\sigma_s$ to demonstrate the model's localization performance under stricter conditions.

The primary localization metrics employed in this paper include precision, recall, and F1-score. Precision assesses the proportion of predicted positive samples that are true positive samples, while recall evaluates the proportion of all true positive samples that are successfully detected by the model. These metrics can be represented by the following formulas:
\begin{equation}
\label{precision}
Precision=\frac{TP}{TP+FP}
\end{equation}
\begin{equation}
\label{recall}
Recall=\frac{TP}{TP+FN}
\end{equation}
where $TP$ denotes true positives, indicating instances where the model correctly matches predicted points with ground truth points; $FN$ represents false negatives, indicating cases where the model fails to match any predicted points for ground truth points; $FP$ signifies false positives, indicating instances where the points predicted by the model lack corresponding ground truth points.

The F1-score, a measure of model accuracy in binary classification, combines precision and recall into a single metric. It ranges from 0 to 1, with higher scores indicating better performance. The F1-score is calculated using the harmonic mean of precision and recall. Mathematically, it can be expressed as:
\begin{equation}
\label{f1}
F1=\frac{(2\cdot precision \cdot recall)}{(precision + recall)}
\end{equation}
\subsubsection{Counting}
In addition, we use Mean Absolute Error (MAE) and Root Mean Squared Error (RMSE) to evaluate the model’s counting ability, indicating that our model has excellent generalization on other tasks. The definitions of MAE and RMSE are as follows:
\begin{equation}
\label{mae}
MAE=\frac{1}{N} \sum_{i=1}^{N} |y_i - \hat{y}_i| 
\end{equation}
\begin{equation}
\label{rmse}
RMSE=\sqrt{\frac{1}{N} \sum_{i=1}^{N} (y_i - \hat{y}_i)^2}
\end{equation}
where $N$ represents the total number of test images, $y_i$ denotes the number of ground truth points in the i-th image, and $\hat{y}_i$ represents the number of predicted points in the i-th image. Lower values of MAE and RMSE indicate better alignment between the model's predictions and the ground truth results.

\subsection{Implementation Details}
In our experiments, query images are uniformly resized to $512 \times 512$, and the extracted query features are down-sampled to $128 \times 128$. At the same time, support features are uniformly cropped to $3 \times 3$. The query features and support features are then uniformly projected to 256 dimensions. All experiments are implemented on a single NVIDIA RTX 3090. Since we mainly experiments on FSC-147 dataset, while support samples in this dataset are cropped from query images, the batch size is set to 1 to prevent negative impacts from other images. The model uses the Adam optimizer, with a learning rate of 0.00002, which drops by 0.25 every 80 epochs. All experiments are run for 200 epochs, and the epoch with the highest F1-score under $\sigma_l=10$ is taken as the optimal result.

\subsection{Experiment Results and Analysis}
Since this paper introduces the first FSOL framework, there are no existing methods available for comparison. To evaluate the FSOL model's performance, we adapt the current state-of-the-art few-shot counting method, SafeCount \cite{SafeCount}, into a localization model for comparison. Initially, experiments are conducted on the FSC-147 dataset, revealing superior performance of our model compared to the most advanced few-shot counting method, SafeCount (block = 1), while also significantly reducing computational resources and training time. Furthermore, localization performance tests are carried out on three dense datasets to further validate the model's performance. Even when compared with robust supervised models, our model demonstrates competitive performance under the few-shot setting. Additionally, ablation experiments are conducted to verify the effectiveness and necessity of the DFA and SQ modules. These findings suggest that our model exhibits strong generalization ability and potential, establishing a solid baseline for the field of FSOL.

To assess the counting and localization performance of the FSOL model, we compare it with SafeCount on the FSC-147 dataset. Initially, we integrate SafeCount into our pipeline as a localization model, testing it with both 4 and 1 SafeCount blocks \textcolor{black}{under 1-shot setting.} We then record the computational consumption of our FSOL model, the baseline model (without DFA and SQ modules), and SafeCount models with varying block counts, including parameters, GFLOPs, and average processing time per image. For fairness, we average results over ten experiments, especially for processing time per image, where we calculate the average time required for the model to process 100 images across ten experiments. Furthermore, we demonstrate the performances of our FSOL model in both 2-shot and 3-shot settings to showcase its generalization capabilities, while also highlighting its ability to prevent overfitting in the 1-shot setting. Table \ref{fsc} summarizes these findings.

%moved here for pagination purposes
\begin{table*}[!t]
\begin{center}
\caption{Quantitative comparison results on FSC-147 dataset. The comparison metrics encompass the model's computational cost, as well as its localization and counting performance.}
\label{fsc}
\begin{tabular}{c c c c c c c c}
\toprule
\multirow{3}{*}{Methods} & \multicolumn{3}{c}{Computational Cost} & \multicolumn{2}{c}{Counting} & \multicolumn{2}{c}{Localization} \\
\cmidrule(lr){2-4}\cmidrule(lr){5-6}\cmidrule(lr){7-8}
&Params(M)$\downarrow$ & GFLOPs(G)$\downarrow$ & Speed(ms)$\downarrow$ & MAE$\downarrow$ & MSE$\downarrow$ & F1($\sigma=5$)$\uparrow$ & F1($\sigma=10$)$\uparrow$\\
\midrule
SafeCount(1 block)&\textbf{53.22}&\textbf{474.86}&\textbf{184.30}&24.4&70.84&52.76&67.94 \\
SafeCount(4 blocks)&67.78&713.37&297.25&\textbf{14.36}&\textbf{58.46}&\textbf{59.48}&\textbf{75.42} \\
\midrule
Baseline&\textbf{48.44}&\textbf{113.59}&\textbf{27.00}&23.05&77.11&43.41&62.57 \\
1-shot FSOL&48.57&117.38&60.40&20.33&68.28&53.40&70.00 \\
2-shot FSOL&48.57&118.60&85.27&20.78&72.29&54.15&70.59 \\
3-shot FSOL&48.57&119.82&105.74&\textbf{19.85}&\textbf{68.25}&\textbf{54.75}&\textbf{71.36} \\
\bottomrule
\end{tabular}
\end{center}
\end{table*}
The results demonstrate that the FSOL model maintains competitive localization and counting performance compared to the SafeCount model, while significantly reducing the required parameter quantity, computational resources, and computation speed. \textcolor{black}{SafeCount utilizes transformer-based multi-head self-attention and direct query feature addition to achieve weighted feature aggregation and residual feature fusion. Nonetheless, within our FSOL pipeline, the weighted feature aggregation is substituted with 3D convolution of information-enhanced features, while residual fusion is achieved through selective Self-Query weighting guided by the query featu re. Consequently, our FSOL pipeline not only achieves superior localization performance but also reduces the model's complexity.} Specifically, compared to the 1-block and 4-block SafeCount models, the FSOL model reduces the parameter quantity by approximately 5M and 20M, respectively. The GFLOPs are approximately one-fifth and one-seventh, respectively, and the computation speed is approximately half and one-fourth, respectively. Under these conditions, when $\sigma_l=10$, the F1-score increases by 2\% and decreases by 5\%, respectively. In such a trade-off, the FSOL model not only improves both counting and localization performance compared to the 1-block SafeCount model but also saves computational resources, highlighting the excellent performance and potential of the FSOL model.

\begin{figure*}[!tp]
\centering
{\includegraphics[width=7in]{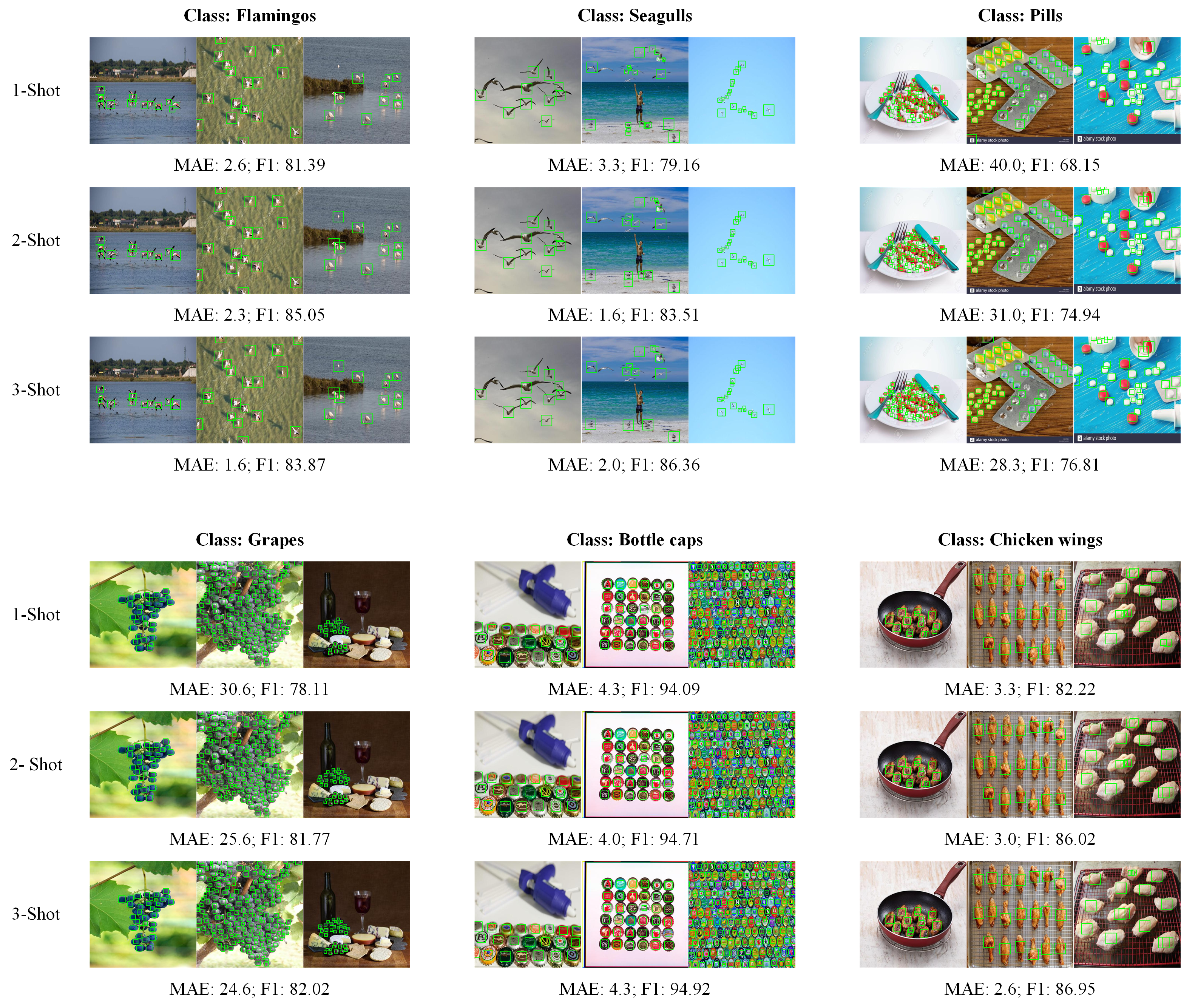}}%
\caption{\textcolor{black}{Qualitative localization results on the FSC-147 dataset are visualized. The MAE and F1-score represent the mean values calculated from three images under various experimental settings. The FSOL model achieves a lower MAE and a higher F1-score when provided with an increased number of support samples.}}
\label{Visual}
\end{figure*}

\textcolor{black}{Furthermore, the counting and localization performances of our FSOL model in 1-shot, 2-shot, and 3-shot settings underscore its robustness in few-shot scenarios. In other words, our FSOL model is capable of effectively learning from the information within the support samples and generalizing to novel samples, even when the number of provided support samples is limited. Moreover, our FSOL model achieves its optimal performance in the 3-shot setting. This suggests that our FSOL model can exploit an increased number of given samples to enhance its localization capabilities, while also demonstrating superior adaptability, thereby achieving competitive performance across various experimental settings.} \textcolor{black}{Further, we visualize the localization results of our FSOL model on the FSC-147 dataset, as depicted in Fig. \ref{Visual}. Through qualitative analysis, it is evident that our FSOL model is capable of handling both dense and sparse objects with competitive performance. Additionally, the multi-shot results underscore the learning ability and robustness of our FSOL model.}

To investigate the model’s performance in dense object localization tasks, we conduct experiments on dense datasets ShangHaiTech PartA, ShangHaiTech PartB, and CARPK \cite{ShangHaiTech, CARPK}. Additionally, we record the performance of the baseline model to assess the performance improvement brought by the two modules. We select five strong supervised localization models: FIDTM \cite{FIDTM}, CLTR \cite{CLTR}, GMS \cite{GMS}, IIM \cite{IIM}, and the STEERER \cite{STE} for comparison with the FSOL model and Baseline model. This helps demonstrate the significant localization performance improvement brought by our model. FIDTM and CLTR use a low threshold $\sigma_s$ of 4 and a high threshold $\sigma_l$ of 8. In this experiment, the FSOL model uses thresholds consistent with these. GMS, IIM, and STEERER do not provide low threshold results; they use a high threshold of $\sigma_l = \sqrt{w^2+h^2}/2$, where $w$ and $h$ represent the width and height of the sample. For consistency with FSOL, we choose point annotation model results for IIM during comparison. The results in Table \ref{sa} and Table \ref{sb} show that the FSOL model demonstrates competitive performance compared to strong supervised models, even in few-shot settings.

%moved here for pagination purposes
\begin{table}[!t]
\begin{center}
\caption{Quantitative comparison results on ShangHaiTech A dataset.}
\label{sa}
\begin{tabular}{c c c c c c c c}
\toprule
\multirow{3}{*}{Supervised} & \multirow{3}{*}{Methods} & \multicolumn{3}{c}{Threshold=$\sigma_s$} & \multicolumn{3}{c}{Threshold=$\sigma_l$} \\
\cmidrule(lr){3-5}\cmidrule(lr){6-8}
& & F1 & AP & AR & F1 & AP & AR \\
\midrule
\multirow{5}{*}{Full} & IIM(point)&-&-&-&73.3&76.3&70.5 \\
&CLTR&43.2&43.6&42.7&74.2&74.9&73.5\\
&FIDTM&\textbf{58.6}&\textbf{59.1}&\textbf{58.2}&77.6&78.2&77.0\\
&GMS&-&-&-&78.1&\textbf{81.7}&74.9\\
&STEERER&-&-&-&\textbf{79.8}&80.0&\textbf{79.4}\\
\midrule
\multirow{2}{*}{1-shot} & Baseline&37.7&39.9&35.7&60.6&64.2&57.5 \\
&FSOL&\textbf{52.4}&\textbf{58.4}&\textbf{47.6}&\textbf{69.6}&\textbf{77.6}&\textbf{63.1}\\
\bottomrule
\end{tabular}
\end{center}
\end{table}

%moved here for pagination purposes
\begin{table}[!t]
\begin{center}
\caption{Quantitative comparison results on ShangHaiTech B dataset.}
\label{sb}
\begin{tabular}{c c c c c c c c}
\toprule
\multirow{3}{*}{Supervised} & \multirow{3}{*}{Methods} & \multicolumn{3}{c}{Threshold=$\sigma_s$} & \multicolumn{3}{c}{Threshold=$\sigma_l$} \\
\cmidrule(lr){3-5}\cmidrule(lr){6-8}
& & F1 & AP & AR & F1 & AP & AR \\
\midrule
\multirow{4}{*}{Full} & FIDTM&\textbf{64.7}&\textbf{64.9}&\textbf{64.5}&83.5&83.9&83.2 \\
&IIM(point)&-&-&-&83.8&89.8&78.6\\
&GMS&-&-&-&86.3&\textbf{91.9}&81.2\\
&STEERER&-&-&-&\textbf{87.0}&89.4&\textbf{84.8}\\
\midrule
\multirow{2}{*}{1-shot} & Baseline&56.6&60.4&53.0&73.0&78.0&68.6 \\
&FSOL&\textbf{67.2}&\textbf{75.5}&\textbf{60.5}&\textbf{78.0}&\textbf{88.4}&\textbf{70.9}\\
\bottomrule
\end{tabular}
\end{center}
\end{table}

We also conduct model tests on the remote sensing dataset CARPK, using experimental settings of $\sigma_s=5$ and $\sigma_l=10$. From the experimental results in Table \ref{carpk}, it can be seen that when $\sigma_l=10$, the F1-score increases from 84.76\% to 93.46\%. This indicates that the DFA and SQ module greatly improve the overall performance.

%moved here for pagination purposes
\begin{table}[!t]
\begin{center}
\caption{Quantitative comparison results on CARPK dataset.}
\label{carpk}
\begin{tabular}{c c c c c c c}
\toprule
\multirow{3}{*}{Methods} & \multicolumn{3}{c}{Threshold=$\sigma_s$} & \multicolumn{3}{c}{Threshold=$\sigma_l$} \\
\cmidrule(lr){2-4}\cmidrule(lr){5-7}
& F1 & AP & AR & F1 & AP & AR \\
\midrule
Baseline&64.35&62.05&66.81&84.76&81.74&88.01\\
FSOL&\textbf{81.84}&\textbf{80.90}&\textbf{82.80}&\textbf{93.46}&\textbf{92.38}&\textbf{94.56}\\
\bottomrule
\end{tabular}
\end{center}
\end{table}

\subsection{Ablation Study}
To verify the effectiveness of our modules, we conduct extensive ablation studies on the FSC-147 dataset. We experiment by removing the Self Query (SQ) module, Dual-path Feature Augmentation (DFA) module, Deformable Convolution (DC) used within DFA, and Cross Central Difference Convolution (CCD-C) used in DFA, respectively. Table \ref{sq} records the quantitative analysis of the SQ experiment results, while Table \ref{dfa} presents the quantitative analysis of the remaining experiments. The results indicate that these modules have a positive impact on localization performance.

\subsubsection{Self Query Ablation}
To illustrate the effectiveness of the SQ module, we remove it and compare the experimental results with the FSOL model. Table \ref{sq} demonstrates that removing SQ notably decreases the localization performance on both the validation set and the test set. Specifically, when the low threshold $\sigma_s=5$ is selected, indicating stricter judgment of positive and negative samples, the F1-score decreases by 2.86\% and 2.07\% respectively on the validation and test sets. Similarly, the AP decreases by 4.51\% and 2.51\% respectively, and AR decreases by 1.35\% and 1.55\% respectively. When the high threshold $\sigma_l=10$ is used, the F1-score decreases by 2.79\% and 2.66\% respectively on the validation and test sets. The AP decreases by 4.94\% and 3.31\% respectively, and AR decreases by 0.83\% and 1.89\% respectively. This underscores the significant contribution of the SQ module to FSOL.

%moved here for pagination purposes
\begin{table*}[!t]
\begin{center}
\caption{Quantitative comparison results between FSOL and SQ-removed models.}
\label{sq}
\begin{tabular}{c c c c c c c c c c c c c}
\toprule
\multirow{5}{*}{Self query} & \multicolumn{6}{c}{Validation Set} & \multicolumn{6}{c}{Test Set} \\
\cmidrule(lr){2-7}\cmidrule(lr){8-13}
& \multicolumn{3}{c}{Threshold=$\sigma_s$} & \multicolumn{3}{c}{Threshold=$\sigma_l$} & \multicolumn{3}{c}{Threshold=$\sigma_s$} & \multicolumn{3}{c}{Threshold=$\sigma_l$}\\
\cmidrule(lr){2-4}\cmidrule(lr){5-7}\cmidrule(lr){8-10}\cmidrule(lr){11-13}
& F1 & AP & AR & F1 & AP & AR & F1 & AP & AR & F1 & AP & AR\\
\midrule
$\times$&50.5&51.0&50.1&67.2&67.9&66.6&46.8&44.7&49.0&68.1&65.0&71.5 \\
$\checkmark$&53.4&55.5&51.4&70.0&72.8&67.4&48.9&47.2&50.6&70.7&68.3&73.4\\
\bottomrule
\end{tabular}
\end{center}
\end{table*}

%moved here for pagination purposes
\begin{table*}[!t]
\begin{center}
\caption{Quantitative comparison results among FSOL, DFA-removed and CCD-C or DC replaced models.}
\label{dfa}
\begin{tabular}{c c c c c c c c c c c c c c}
\toprule
\multicolumn{2}{c}{\multirow{3}{*}{DFA}} & \multicolumn{6}{c}{Validation Set} & \multicolumn{6}{c}{Test Set} \\
\cmidrule(lr){3-8}\cmidrule(lr){9-14}
 & & \multicolumn{3}{c}{Threshold=$\sigma_s$} & \multicolumn{3}{c}{Threshold=$\sigma_l$} & \multicolumn{3}{c}{Threshold=$\sigma_s$} & \multicolumn{3}{c}{Threshold=$\sigma_l$}\\
\cmidrule(lr){1-2}\cmidrule(lr){3-5}\cmidrule(lr){6-8}\cmidrule(lr){9-11}\cmidrule(lr){12-14}
CCD-C & DC & F1 & AP & AR & F1 & AP & AR & F1 & AP & AR & F1 & AP & AR\\
\midrule
$\times$&$\times$&47.2&49.6&45.0&64.8&68.0&61.8&43.2&41.8&44.8&66.0&63.8&68.3\\
$\times$&$\checkmark$&52.3&54.1&50.5&69.3&71.7&67.0&47.8&45.8&50.0&69.9&67.0&73.1\\
$\checkmark$&$\times$&51.4&53.4&49.6&68.3&70.9&65.8&46.3&44.3&48.4&67.7&64.9&70.8\\
$\checkmark$&$\checkmark$&53.4&55.5&51.4&70.0&72.8&67.4&48.9&47.2&50.7&70.7&68.3&73.4\\
\bottomrule
\end{tabular}
\end{center}
\end{table*}

\subsubsection{Dual-path Feature Augmentation Ablation}
To demonstrate the effectiveness of the DFA module, we uniformly replace Deformable Convolution (DC) and Cross Central Difference Convolution (CCD-C) with vanilla 2D convolution. This ensures that the query feature $F_Q\in\mathbb{R}^{1\times256\times128\times128}$ and support feature $F_S\in\mathbb{R}^{1\times256\times3\times3}$ are consistent in spatial dimensions with the output features of the DFA module. Table \ref{dfa} shows that when the DFA module is removed or modified with vanilla convolution, the localization performance significantly decreases on both the validation set and the test set. We focus on the performance change when F1-score on the validation set and test set when the high threshold $\sigma_l=10$ is selected. Specifically, when the DFA module is removed, that is, the CCD-C and DC used for feature enhancement are all replaced with vanilla convolution, the F1-score decreases by 5.25\% and 4.76\% respectively; when the CCD-C in DFA is replaced with vanilla convolution, the F1-score decreases by 0.74\% and 0.83\% respectively; as for DC is replaced, the F1-score decreases by 1.74\% and 3.02\% respectively. This shows that the DFA module not only stimulates the potential of CCD-C convolution and DC in the two branches respectively, but also positively contributes to the feature fusion of the two branches.

\begin{figure*}[!tp]
\centering
{\includegraphics[width=7in]{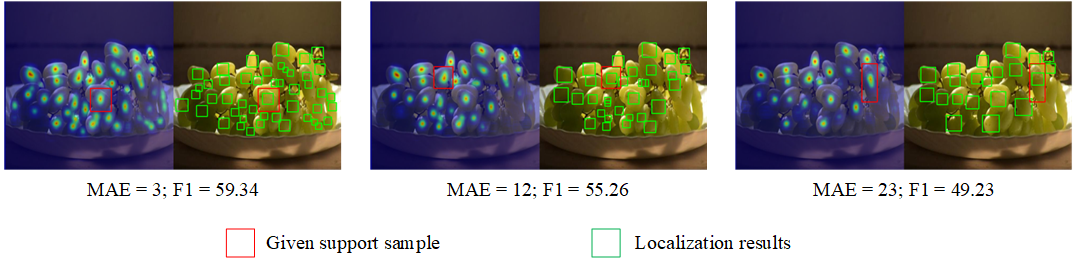}%
\caption{\textcolor{black}{Visualization of the localization results under different support samples. The figure showcases various selections of support samples and visualizes the corresponding localization results. It is clear that the choice of support samples greatly affects the model's localization performance.}}
\label{Diff}}
\end{figure*}

\section{Future Work}
Our FSOL framework establishes a strong baseline for the FSOL task, providing a solid foundation for future research in this field. While our Dual-path Feature Augmentation (DFA) and Self Query (SQ) modules have demonstrated competitive performance enhancements, their simplicity suggests potential for further refinement. In future work, we aim to improve our approach in the following ways:
% \subsubsection{SQ Module Upgrade}
% Building upon the self query concept, we aim to design adaptive structures to enhance the FSOL framework's compatibility with diverse styles of query images. This will leverage the self query capabilities of the SQ model, empowering the FSOL framework with cross-domain matching capabilities.
% \subsubsection{DFA Module Upgrade}
% The current DFA module operates with two branches to enhance deformation and gradient information, subsequently fusing these aspects using 3D convolution. Future endeavors could delve into extracting and incorporating additional feature information conducive to sample matching, and devise more interpretable and elegant fusion strategies.
\subsubsection{\textcolor{black}{Support Samples Selection}}
\textcolor{black}{In our 1-shot FSOL pipeline, the selection of the support sample from the available dataset significantly impacts localization performance. Fig. \ref{Diff} illustrates varying object localization outcomes based on different selected support samples. Looking ahead, our goal is to develop a more adaptive and robust localization method capable of extracting deeper semantic information from the provided support samples, thereby ensuring precise and reliable localization.}
\subsubsection{Model Architecture Upgrade}
Future research could delve into exploring more advanced architectures, such as Transformer or the latest Mamba architecture \cite{Mamba}. These architectures hold promise in capturing intricate patterns and dependencies within the data, potentially leading to improved accuracy in object localization.
\subsubsection{Task Upgrade}
In future research, there could be an emphasis on diversifying the modalities of query prompts. For example, exploring the use of natural language instead of support prompt images to search for corresponding information in query images. This approach could potentially facilitate the transition from few-shot object localization to zero-shot object localization, thereby making significant contributions to a broader field.

\section{Conclusion}
This paper defines the FSOL task for the first time and proposes a high-performance baseline framework. The framework consists of two main modules: The DFA module ingeniously strengthens the association between support samples and query images through its dual-branch design, mitigating significant intra-class appearance gaps and addressing object omission challenges; The SQ module integrates the distribution information of the query image into the similarity map, resulting in a substantial enhancement of the model's performance. Experimental results on both sparse and dense datasets demonstrate the robust performance and significant application potential of our FSOL model. With this high-performance FSOL framework, the model can effectively handle unseen target categories given only a few labeled samples, greatly facilitating future studies in FSOL task as well as expanding the scope of object localization applications.

\section*{Acknowledgment}
The research project is partially supported by National Key R\&D Program of China (No. 2021ZD0111902), NSFC(No. 62072015, U21B2038).

\normalem
\bibliographystyle{IEEEtran}
\bibliography{FSOL}{}

\end{document}